\documentclass[letterpaper]{article} % DO NOT CHANGE THIS
\usepackage{aaai25}  % DO NOT CHANGE THIS
\usepackage{times}  % DO NOT CHANGE THIS
\usepackage{helvet}  % DO NOT CHANGE THIS
\usepackage{courier}  % DO NOT CHANGE THIS
\usepackage[hyphens]{url}  % DO NOT CHANGE THIS
\usepackage{graphicx} % DO NOT CHANGE THIS
\usepackage{natbib}  % DO NOT CHANGE THIS AND DO NOT ADD ANY OPTIONS TO IT
\usepackage{caption} % DO NOT CHANGE THIS AND DO NOT ADD ANY OPTIONS TO IT
\usepackage[ruled,plain,lined,shortend,linesnumbered]{algorithm2e}
\usepackage{algorithmic}

\usepackage{multirow}
\usepackage{multicol}
\usepackage{booktabs}
\usepackage{float}

\usepackage{amsmath}
\usepackage{amsfonts}
\usepackage{appendix}
\usepackage{stfloats}
\usepackage{afterpage}

\usepackage{newfloat}
\usepackage{listings}
\DeclareCaptionStyle{ruled}{labelfont=normalfont,labelsep=colon,strut=off} % DO NOT CHANGE THIS
\floatstyle{ruled}
\newfloat{listing}{tb}{lst}{}
\floatname{listing}{Listing}
\title{FedCFA: Alleviating Simpson's Paradox in Model Aggregation \\ with Counterfactual Federated Learning}
\author{
    Zhonghua Jiang\textsuperscript{\rm 1}\equalcontrib,
    Jimin Xu\textsuperscript{\rm 1}\equalcontrib, 
    Shengyu Zhang\textsuperscript{\rm 1}\thanks{The corresponding author.}, 
    Tao Shen\textsuperscript{\rm 1}, \\
    Jiwei Li\textsuperscript{\rm 1}, 
    Kun Kuang\textsuperscript{\rm 1}, 
    Haibin Cai\textsuperscript{\rm 2}, 
    Fei Wu\textsuperscript{\rm 1} 
}
\affiliations{
    \textsuperscript{\rm 1}Zhejiang University\\
    \textsuperscript{\rm 2}East China Normal University\\
    \{jiangzhonghua, xujimin, sy\_zhang, tao.shen, jiwei\_li, kunkuang, wufei\}@zju.edu.cn,
    hbcai@sei.ecnu.edu.cn
}

\usepackage{bibentry}
\begin{document}

\maketitle

\begin{abstract}
Federated learning (FL) is a promising technology for data privacy and distributed optimization, but it suffers from data imbalance and heterogeneity among clients. Existing FL methods try to solve the problems by aligning client with server model or by correcting client model with control variables. These methods excel on IID and general Non-IID data but perform mediocrely in Simpson's Paradox scenarios. Simpson's Paradox refers to the phenomenon that the trend observed on the global dataset disappears or reverses on a subset, which may lead to the fact that global model obtained through aggregation in FL does not accurately reflect the distribution of global data. Thus, we propose FedCFA, a novel FL framework employing counterfactual learning to generate counterfactual samples by replacing local data critical factors with global average data, aligning local data distributions with the global and mitigating Simpson's Paradox effects. In addition, to improve the quality of counterfactual samples, we introduce factor decorrelation (FDC) loss to reduce the correlation among features and thus improve the independence of extracted factors. We conduct extensive experiments on six datasets and verify that our method outperforms other FL methods in terms of efficiency and global model accuracy under limited communication rounds.
\end{abstract}


\section{Introduction}
Federated learning (FL) is a technology to enable collaborative learning among multiple parties without violating data privacy \cite{konevcny2016federated}. 
In FL, many clients collaborate to train a shared model under the coordination of a server while keeping data dispersed. This reduces the risk of privacy breaches generated by centralized machine learning.

FedAvg \cite{mcmahan2017communication}, a basic FL algorithm, applies gradient descent to train models on distributed clients. 
In order to adapt to the special scenarios of FL, many researchers have proposed various improved algorithms \cite{reguieg2023comparative,li2021ditto,rothchild2020fetchsgd} based on FedAvg to enhance the fairness of resource allocation \cite{ilhan2023scalefl, hao2021towards}, communication efficiency \cite{liao2023adaptive, zhao2023resource}, privacy security \cite{zhang2021survey, li2022auditing} and defense attack ability \cite{ovi2023mixed, cao2022mpaf} of FL. Among them, the imbalance and heterogeneity \cite{luo2023gradma,mendieta2022local,qu2022rethinking} of client data are one of the most prominent problems in FL.

\begin{figure}[t]
    \centering
       \includegraphics[width=1.0\linewidth]{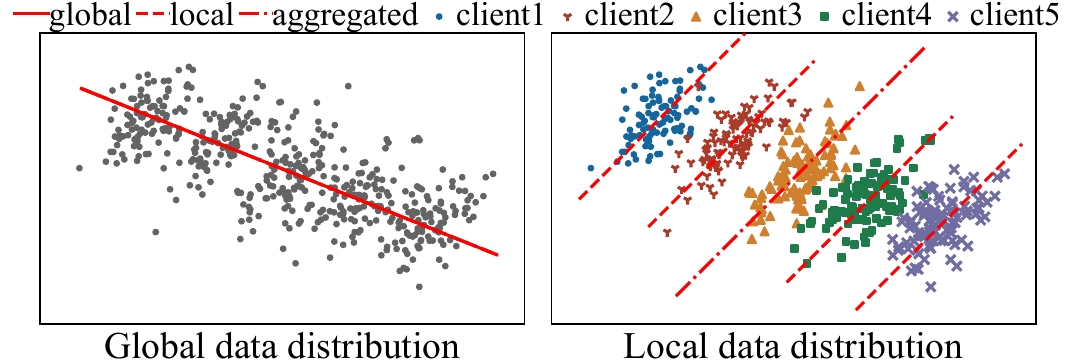}
       \caption{Simpson's Paradox.}
       \label{fig:Simpson_paradox}
% \end{figure}
% \begin{figure}[t]
    \centering
       \includegraphics[width=1.0\linewidth]{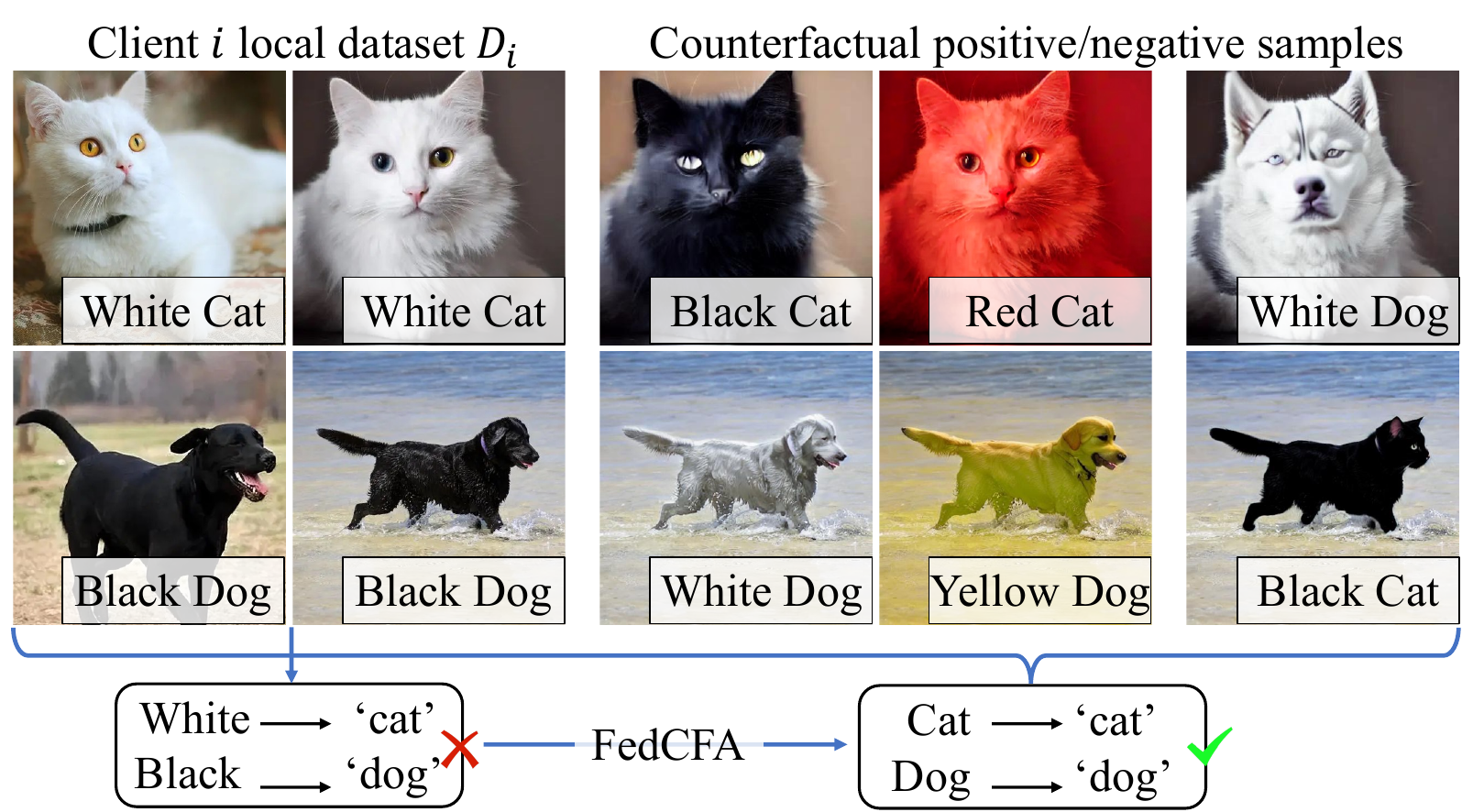}
       \caption{FedCFA can generate counterfactual samples that do not exist locally on client $i$, preventing the model from learning incorrect feature-label relationships.}
       \label{fig:fedcfa_data}
\end{figure}

The inherent variability in data across clients presents a formidable challenge to FL, causing locally-trained client models to overlook broader patterns evident in the global data. Such heterogeneity further results in inefficient collaborative training and increased communication rounds. To address this problem, existing FL methods mainly adopt two strategies. The first involves \cite{fang2022robust,seo2023madfed,li2023synthetic} leveraging data outside of local data to enhance local model optimization, albeit at the potential cost of compromising data privacy. The second strategy \cite{cheng2022differentially,gao2022feddc,li2021model} involves heuristically adjusting local models to align more closely with the global model, thereby alleviating data heterogeneity. While most existing works leverage the global model as the alignment reference, our observations, as illustrated in Figure \ref{fig:Simpson_paradox}, indicate that \textbf{global model may fall victim to Simpson's Paradox and potentially become untrustworthy}. That is, the data distribution captured by the aggregated model contradicts the global data distribution, rendering the aforementioned strategies ineffective.  

Simpson's Paradox manifests in probability and statistics as a discrepancy where a trend in a dataset reverses or disappears upon analyzing subsets or aggregated data, as depicted in Figure \ref{fig:Simpson_paradox}.
Consider a FL system for classifying cat and dog images, involving two clients with distinct datasets. Client $i$'s dataset primarily includes images of white cats and black dogs, and Client $j$'s dataset comprises images of light gray cats and brown dogs. Individually, the datasets reveal a similar trend: lighter-colored animals are categorized as 'cat', while darker-colored animals are deemed 'dog'. This leads in an aggregated global model that leans towards associating colors with classifications and assigning higher weights to color features. Nevertheless, the global data distribution introduces a number of images of cats and dogs with different colors (such as black cats and white dogs), contradicting the aggregated model. A model trained on global data can easily discover that animal colors are not related to specific classifications, thus reducing weights of color features.

In this paper, we set the goal of alleviating Simpson's Paradox problem, \textit{i.e.}, aligning the aggregated model with global data distribution.
We propose FedCFA, a novel FL framework leveraging counterfactual learning.
The essence of FedCFA in confronting Simpson's Paradox problem is to identify and counterfactually manipulate critical features that dominate the deviation from global data distribution under the guidance of globally aggregated data, to make local data distribution closer to global data distribution.
As shown in Figure \ref{fig:fedcfa_data}, the counterfactual samples generated through the counterfactual transformation enable the local model to grasp accurate feature-label relationship and avoid local data distribution contradicting global data distribution, thereby alleviating Simpson's Paradox in model aggregation.
Technically, we devise two counterfactual modules that selectively replace critical features, integrating global average data into local data, and constructs positive/negative samples for model learning.
Specifically, given local data, we identify dispensable/indispensable features, performing counterfactual transformations to obtain positive/negative samples by replacing those features accordingly.
Through contrastive learning on counterfactual samples which are \textbf{closer to the global data distribution}, the local model can effectively learn global data distribution. 
However, counterfactual transformation faces the challenge of extracting independently controllable features from data. A feature may encode multiple types of information, e.g., a pixel of an animal image may carry both color and shape information. To improve the quality of counterfactual samples, we need to ensure that the extracted features cover singular information. Therefore, we introduce factor decorrelation loss, which directly punishes the correlation coefficient between factors to achieve decoupling between feature.

We conduct in-depth experiments to verify the effectiveness of FedCFA on different datasets.
To summarize, this paper makes the following key contributions:
\begin{itemize}
    \item We point out the limitations of existing FL algorithms in dealing with Simpson's Paradox and propose a new FL framework based on counterfactual learning, namely FedCFA, for mitigating this problem.
    % which can solve this problem to a certain extent.
    \item Based on FedCFA framework, we design and implement a baseline. This baseline can decouple the data features, obtain independent factors for counterfactual transformation, generates counterfactual samples for local model training, and avoids learning wrong data distribution. Finally, we design a factor decorrelation loss function to measure and constrain the correlation among factors, enhancing feature decoupling effectiveness.
    \item We conduct extensive experiments based on six datasets under Non-IID and IID data distribution settings. Compared with FedAvg and the advanced FL algorithms, the proposed FedCFA has better convergence and training efficiency, improving the accuracy of the global model.
\end{itemize}

\section{Related Works}
FL is a distributed machine learning paradigm that allows multiple clients to collaboratively train a shared model while protecting data privacy \cite{konevcny2016federated}. 
FedAvg \cite{mcmahan2017communication} is a widely used FL algorithm that achieves model sharing by performing local gradient descent on clients and averaging parameters on server.
% This algorithm can handle general distributed optimization problems well. 
To adapt to FL's special scenarios, many improved algorithms \cite{reguieg2023comparative,li2021ditto,zhu2023confidence,rothchild2020fetchsgd,li2021fedmask,chow2023stdlens,ilhan2023scalefl} have also been proposed based on FedAvg. 
Among these, training a high-quality global model amidst data heterogeneity \cite{liTi2020federated, zhu2021federated,xie2023fedkl, hamman2024demystifying} remains a significant FL challenge.

In order to solve the problem of data heterogeneity, some methods \cite{chen2023federated,chai2022cross} try to use global or other clients' data to aid local model training. FedBGVS \cite{mou2021optimized} mitigates class bias impact through a balanced global validation set. CDFDM \cite{chai2022cross} designs a sharing model that dynamically allocates the amount of shared data according to user data scale, effectively alleviating the problem of data heterogeneity. However, FL methods based on shared data will inevitably leak the privacy information of data owner. 
Other methods \cite{li2020federated,karimireddy2020scaffold,li2019fair} adjust local models or aggregate weights for consistency with the global model. FedProx \cite{li2020federated} introduces the difference between the global model and the local model of the previous round as a regularization term in the objective function, so that the local update will not deviate too much from the global model. SCAFFOLD \cite{karimireddy2020scaffold} uses control variables (variance reduction) to correct errors in local updates caused by client drift. 
q-FedAvg \cite{li2019fair} considers the issue of federation fairness and reduces the variance by adjusting aggregation weight, albeit at the cost of diminished model performance on some clients.
However, when Simpson's Paradox exists in data, neither client model nor global model can accurately reflect the true distribution of global data, causing the above strategy to fail. 
FedMix \cite{yoon2021fedmix} augments local data via Mixup with global average data. It fails to disrupt spurious feature-label relationship in local data, and still cannot mitigate Simpson's Paradox.
Counterfactual learning \cite{zhang2024transferring,alfeo2023local,zhang2020devlbert} partly mitigates Simpson's Paradox by generating global-distribution-aligned samples through local data intervention.

\section{Method}
\subsection{Problem Statement}
We consider a FL scenario with K clients, each holding a local dataset $\mathcal{D}_{k}$, for $k=1,2,...,K$. Let $\mathcal{D}_{g}=\cup^{K}_{k=1}\mathcal{D}_{k}$ denote the global dataset, and let $|\mathcal{D}|$ denote the size of dataset $\mathcal{D}$. The primary objective in FL is to aggregate model parameters $w_k$ from all participating clients, aiming to derive a global model $w_g$
that fits the global data distribution. This problem is formalized as an optimization task, where $\ell(\cdot)$ denotes the loss function:
\begin{equation}
\begin{split}
\small
  \min_{w_{g}} \{ f_{g}(w_{g}) &:= \frac{1}{|\mathcal{D}_{g}|}\textstyle \sum_{(x,y) \in \mathcal{D}_{g}} \ell(w_{g}; x,y) \}, \\
  s.t. \, \ w_{g}&=\textstyle \sum_{k=1}^{K}\frac{|\mathcal{D}_{k}|}{|\mathcal{D}_{g}|}w_{k}.
\label{eq:minfg}
\end{split}
\end{equation}
\subsection{Overall Schema}
To address the global model optimization in FL, prevalent algorithms such as FedAvg employ parameter aggregation techniques. Specifically, these algorithms train local models on individual datasets and subsequently aggregate these models' parameters via a weighted average to form a global model. This approach fundamentally assumes that optimizing the function below indirectly tackles the global model optimization issue delineated by Eq. \ref{eq:minfg}, where $\textbf{w}=(w_1, w_2,..., w_K)$:
\begin{equation}
\small
  \min_{\textbf{w}} \{ F(\textbf{w}):=\frac{1}{|\mathcal{D}_{g}|} \textstyle \sum_{k=1}^{K} \textstyle \sum_{(x,y) \in \mathcal{D}_{k}}\ell(w_{k};x,y)\}.
\end{equation}

\begin{algorithm}[t]
    \caption{FedCFA.}
    \label{alg:FedCFA}
    \SetAlgoLined
    \DontPrintSemicolon
    Initialize $w_{g}^{0}$ for global server \;
    \For{$t\leftarrow 0$ \KwTo $T-1$}{
        Server sends $w_{g}^{t}$ and $(\bar{X}_{g},\bar{Y}_{g})$ to the clients\;
        $S_{t} \leftarrow K$ clients selected at random\;
        \For{$k \in S_{t}$}{
            $w_{k}^{t} \leftarrow w_{g}^{t}$ \;
            \For{batch$(X,Y) \subseteq (X_{k},Y_{k})$}{
                Calculate $\mathcal{L}_{cls} = \ell(X,Y)$ and $\mathcal{L}_{corr}$\;
                Generate counterfactual samples \;
                Calculate $\mathcal{L}_{pos}, \mathcal{L}_{neg}$ \;
                Calculate $\mathcal{L}_{total}$, $w_{k}^{t} \leftarrow w_{k}^{t} - {\beta}_{t}\bigtriangledown \mathcal{L}_{total}$
            }
            Compute $(\bar{X}_{k},\bar{Y}_{k})$\;
            Send $w_{k}^{t},(\bar{X}_{k},\bar{Y}_{k})$ to server\;
        }
        Update $(\bar{X}_{g},\bar{Y}_{g})$, $w_{g}^{t+1} \leftarrow \sum_{k=1}^{K}\frac{|\mathcal{D}_{k}|}{|\mathcal{D}_{g}|}w_{k}^{t}$\;
    }
\end{algorithm}

The premise for the above assumption to hold is that local data distribution can approximately represent global data distribution. However, in actual FL scenarios, data heterogeneity can lead to Simpson's Paradox. The distribution of global data is different or even opposite to that of a single client. Therefore, a local model that is only optimized on local data cannot accurately reflect global data distribution. Likewise, the global model obtained by aggregating these local models cannot accurately reflect global data distribution. To address this, we use global average data to perform counterfactual transformation on local data to generate counterfactual samples, so that the local data distribution is closer to the global and used for the optimization of the local model.

\begin{figure*}[t]
  \centering
   \includegraphics[width=0.9\linewidth]{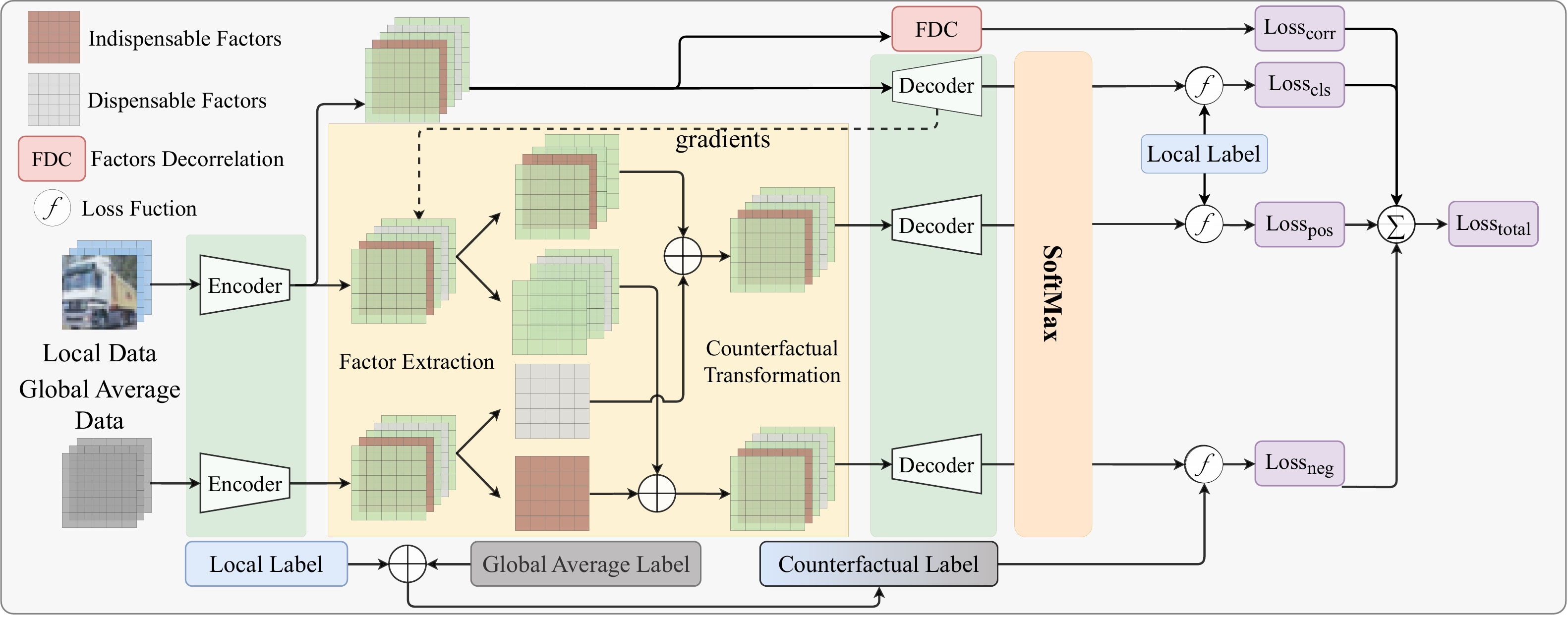}
   \caption{Local model training process in FedCFA.}
   \label{fig:fedcfa}
\end{figure*}

The main steps of our proposed FedCFA framework are shown in Algorithm \ref{alg:FedCFA}, where $T$ is the total communication rounds set for FL, and $w^{t}_{k}$ is the model parameters of client $k$ in the t-th round. 
In FedCFA, we divide a network into three parts: an Encoder for extracting factors, a Decoder for decoding the input factors, and a SoftMax classifier for classifying the features decoded by the decoder.
FedCFA first uses Encoder to extract factors from the image, and then replaces the original factors with factors generated by the global average data through Encoder, thereby generating counterfactual positive and negative samples.
The model performs contrastive learning on counterfactual samples, so that the local model can better fit global data distribution, that is, $w_{k} \Rightarrow w_{r}$, $w_{r}$ represents the right global model parameters. By narrowing the gap between $w_{k}$ and $w_{r}$, it indirectly brings $w_{g}$ and $w_{r}$ closer, that is, $w_{g} \Rightarrow w_{r}$. 
In addition, we propose a factor decorrelation loss to measure the correlation between factors output by Encoder and take it as one of the optimization goals. Through such a design, we can make the factors extracted more independent, thereby improving the quality of counterfactual samples.

\subsection{Global Average Dataset Construction}
According to Central Limit Theorem, for a random subset of size n from the original dataset, the mean $\bar{x}_i$ converges to a normal distribution as n grows large, characterized by a mean $\mu$ and variance $\frac{\sigma^2}{n}$:
\begin{equation}
\small
  \bar{x}_i\sim N(\mu, \frac{\sigma^2}{n}),
  \label{eq:xi-n}
\end{equation}
where $\mu$ and $\sigma^2$ are the expectation and variance of the original dataset. When n is small, $\bar{x}_1$can capture the local characteristics and changes of the dataset more finely, especially in preserving the details of the tail of the data distribution and near the outliers. On the contrary, as n increases, the stability of $\bar{x}_1$ is significantly improved, and its variance is significantly reduced, making it more robust and reliable as an estimate of the overall mean $\mu$, and its sensitivity to outliers is greatly reduced. In addition, in distributed computing environments such as FL, in order to effectively control communication costs, choosing a larger n as the sample size is considered an optimization strategy.

Based on the above analysis, we construct a global average dataset of size B to approximate global data distribution. 
Specifically, each client randomly divides local data $\mathcal{D}_k$ into B subsets of $n=\left \lfloor {|\mathcal{D}_k|}/{B} \right \rfloor$ samples: $\{x_{(i-1)n+1},x_{(i-1)n+2},…,x_{in}\}, i=1,2,…,B$.
For each subset, calculate its mean $\bar{x}_i$:
\begin{equation}
\small
\bar{x}_i = \frac{1}{n}\textstyle \sum_{j=(i-1)n+1}^{i \cdot n} x_j.
  \label{eq:xi}
\end{equation}

As a result, the client can generate a local average dataset $\{\bar{x}_1, \bar{x}_2, \ldots, \bar{x}_B\}$. 
The server aggregates these local average data from multiple clients and uses the same method to compute a global average dataset $\bar{X}_g=\{\bar{x}_{g,1}, \bar{x}_{g,2}, \ldots, \bar{x}_{g,B}\}$ of size B, which approximates global data distribution.
For label $Y$, we apply the identical strategy to generate global average labels.
Finally, we get the complete global average dataset $\bar{\mathcal{D}}_g=\{\bar{X}_g, \bar{Y}_g\}$.

\subsection{Counterfactual Transformation Modules}
\label{subsec:Counterfactual}
As shown in Figure \ref{fig:fedcfa_data}, counterfactual transformation modules replace key features in local data with the corresponding features of global average data, generating counterfactual positive and negative samples, making local data distribution closer to the global, thereby alleviating Simpson's Paradox.
The local model training process in FedCFA is shown in Figure \ref{fig:fedcfa}. Similar to other algorithms, FedCFA first uses original data $(X,Y)$ to calculate classification loss:
\begin{equation}
\small
  \mathcal{L}_{cls}=\ell(X,Y).
  \label{eq:lcls}
\end{equation}

Then, FedCFA uses Encoder to extract data factors $F=Encoder(X)$. Next, we derive the derivative of each factor at the output layer of the model's Decoder to obtain the gradient value of each factor. According to the gradient value, we select the $topk$ factors with low/high gradients, select the corresponding factors from the features obtained by the global average data through the model's Encoder, and replace them to generate counterfactual positive/negative samples.
We use $Mask_{pos}$/$Mask_{neg}$ to set selected low/high gradient factors to zero to retain the factors we need:
\begin{equation}
\small
  F_{pos}=Mask_{pos}*F+(1-Mask_{pos})*\bar{F}_{g} ,
  \label{eq:xpos}
\end{equation}
\begin{equation}
\small
  F_{neg}=Mask_{neg}*F+(1-Mask_{neg})*\bar{F}_{g} ,
  \label{eq:xneg}
\end{equation}
where $\bar{F}_{g}$ represents the factors obtained from global average data processed by the Encoder.

For positive samples, the labels do not change. For negative samples, a weighted average is used to generate counterfactual labels:
\begin{equation}
\small
  Y_{neg}=\frac{topk}{|F|}*\bar{Y}_{g}+(1-\frac{topk}{|F|})*Y ,
  \label{eq:yneg}
\end{equation}
where $\bar{Y}_{g}$ represents the label of the global average data.

The classification loss obtained by FedCFA using counterfactual positive and negative samples is as follows:
\begin{equation}
\small
  \mathcal{L}_{pos}=\ell(F_{pos},Y), \ \ \mathcal{L}_{neg}=\ell(F_{neg},Y_{neg}).
  \label{eq:lpos}
\end{equation}

\subsection{FDC: Factor Decorrelation}
\label{subsec:CorrelationAnalysis}
The same pixel may contain multiple data features. For example, in animal images, a pixel may carry both color and appearance information.
To enable Encoder to efficiently disentangle various factors, we propose a new loss function.
We use Pearson correlation analysis to measure the correlation between factors and use it as a regularization term. Given a batch of data, we use $F_{i}$ to represent all factors of the i-th sample. 
$F_{i,j}$ represents the j-th factor of the i-th sample. We regard the factors of the same index j for each sample in the same batch as a set of variables $F_{;,j}$.
Finally, we use the average of the absolute values of the Pearson correlation coefficients for each pair of variables as the FDC loss:
\begin{equation}
\small
% \mathcal{L}_{corr}=\frac{\sum_{j=1}^{|F_0|}\sum_{j'=1}^{|F_0|}|r(F_{;,j},F_{;,j'})|-|F_{0}|}{|F_0|*|F_0|-|F_0|} ,
\mathcal{L}_{corr}=\frac{\sum_{j=1}^{|F_0|}\sum_{j'>j}^{|F_0|}|r(F_{;,j},F_{;,j'})|}{|F_0|*(|F_0|-1)/2} ,
  \label{eq:lcorr}
\end{equation}
\begin{equation}
\small
  r(F_{;,j},F_{;,j'})=\frac{Cov(F_{;,j},F_{;,j'})}{\sqrt{Var(F_{;,j})Var(F_{;,j'})} } ,
  \label{eq:rcorr}
\end{equation}
where $Cov(\cdot)$ is the covariance calculation function, $Var(\cdot)$ is the Variance calculation function. 

The final total loss is given:
\begin{equation}
\small
    \mathcal{L}_{total} = \mathcal{L} _{cls} + \lambda_{neg}\mathcal{L} _{neg} + \lambda_{pos}\mathcal{L} _{pos} + \lambda_{corr}\mathcal{L} _{corr}.
\label{eq:ltotal}
\end{equation}

\subsection{Proof}
We define $F_k, F'_k, F_g, \bar{F}_g$ to represent the factors obtained by the Encoder for the client's local data, local data after counterfactual transformation, global data, and global average data.
In this paper, we need to prove that the distribution of local data after counterfactual transformation is closer to global data distribution, that is, to prove that the distribution $P_{F'_k}$ of the local data factor $F'_k$ is closer to the distribution $P_{F_g}$ of the global data factor $F_g$.

To quantify the difference between two distributions, we use the Wasserstein distance. For any two probability distributions $P$ and $Q$, the Wasserstein distance is defined as:
\begin{equation}
\small
W(P,Q)=inf_{\gamma \in \Gamma (P,Q)} \mathbb{E}_{(x,y)\sim \gamma}[\left \| x-y \right \|],
\label{eq:Wasserstein}
\end{equation}
where $\Gamma (P,Q)$ is the set of all joint probability distributions with $P$ and $Q$ as marginal distributions.

To prove that the counterfactual transformation can reduce the distance between the local factor distribution $P_{F_k}$ and the global factor distribution $P_{F_g}$, we need to verify the following inequality:
\begin{equation}
\small
W(P_{F_g},P_{F'_k})<W(P_{F_g},P_{F_k}).
\end{equation}
Since we construct a global average dataset to approximate the global data distribution, we only need to prove the following relationship:
\begin{equation}
\small
W(P_{\bar{F}_g},P_{F'_k})<W(P_{\bar{F}_g},P_{F_k}).
\end{equation}
The counterfactual transformation generates $F'_k$ by replacing the key factors $S_k$ of the local factor $F_k$ with the corresponding factors in global average data factor $\bar{F}_g$. This means that for any $f_{k,j} \in F_k$, its transformed sample $f'_{k,j}$ has a higher similarity with $\bar{f}_{g,j}$ of the global average data factor $\bar{F}_g$ on the key feature $S_k$.
Therefore, for any $\bar{f}_g \in \bar{F}_g$ and $f'_{k,j} \in F'_k$, we have
\begin{equation}
\small
\left\| \bar{f}_g - f'_{k,j} \right\|  <  \left\| \bar{f}_g - f_{k,j} \right\|.
\end{equation}
From this, we can conclude that when $\gamma \in \Gamma (P_{\bar{F}_g},P_{F'_k})$, it is expected that $\mathbb{E}_{(x,y)\sim \gamma }[\left \| x-y \right \|]$ is smaller:
\begin{equation}
\small
\begin{split}
    W_1 &=inf_{\gamma \in \Gamma(P_{\bar{F}_g},P_{F'_k})}\mathbb{E}_{(x,y)\sim \gamma}[\left \| x-y \right \|] \\
    &<W_0=inf_{\gamma \in \Gamma (P_{\bar{F}_g},P_{F_k})}\mathbb{E}_{(x,y)\sim \gamma}[\left \| x-y \right \|].
\end{split}
\end{equation}

In summary, $W(P_{\bar{F}_g},P_{F'_k})$ is smaller than $W(P_{\bar{F}_g},P_{F_k})$. This indicates that the counterfactual transformation helps to make the local data closer to the global data distribution.

\section{Experiments}
\label{sec:Experiments}
\newcommand{\vpara}[1]{\vspace{0.07in}\noindent\textbf{#1 }}
\vpara{Implementation.} 
Using FedLab \cite{JMLR:v24:22-0440}, we build a typical FL scenario.
Unless specified otherwise, we use MLP, ResNet18 and LSTM as network model, with 60 clients, learning rate $\beta$ of 0.01, one local epoch, batch size of 128, and 500 communication rounds. 
We conduct experiments on a NVIDIA A100 with 40GB memory.

% \textbf{Datasets.} 

\vpara{Datasets:} 
% We conduct experiments on six datasets commonly used in FL: 
CIFAR10, CIFAR100 \cite{krizhevsky2009learning}, Tiny-ImageNet, FEMNIST \cite{caldas2018leaf}, Sent140 \cite{go2009twitter}, MNIST.
We built a dataset with Simpson's Paradox based on MNIST.

\vpara{Different Data Partition Methods.} We use two different data partition methods: IID and Non-IID. 
IID Partition distributes samples uniformly to K clients through random sampling. 
We use $IID_K$ to represent this data division. %, where $K$ is the number of clients. 
For Non-IID, we utilize Dirichlet distribution $Dir_{K}(\alpha)$ to simulate the imbalance of dataset.
The smaller the $\alpha$, the greater the data difference between clients.
We try several different client numbers and data partition methods: $Dir_{60}(0.6)$, $Dir_{60}(0.2)$, $Dir_{100}(0.2)$, $Dir_{100}(0.6)$, $IID_{60}$, $IID_{100}$.
We use Dirichlet distribution to adjust the frequency of different categories labels in each client to simulate label distribution $P(Y)$ heterogeneity among clients. 
For FEMNIST, we divide different users into different clients to simulate feature distribution $P(X)$ heterogeneity due to handwriting style variance.
For binary classification text dataset Sent140, we divide it into different clients based on users and ensure consistent label distribution among clients, to simulate the heterogeneity of conditional feature distribution $P(X|Y)$.	

\vpara{Baseline Methods:} 
FedAvg \cite{mcmahan2017communication}, FedProx \cite{li2020federated}, SCAFFOLD \cite{karimireddy2020scaffold}, FedPVR \cite{li2023effectiveness}, q-FedAvg \cite{li2019fair} and FedMix \cite{yoon2021fedmix}.

\begin{table*}[t]
\footnotesize
\centering
\setlength{\tabcolsep}{1em}
\begin{tabular}{cccccccc}
\toprule
                         & Method                  &$Dir_{60}(0.2)$ &$Dir_{60}(0.6)$ & $IID_{60}$      &$Dir_{100}(0.2)$ &$Dir_{100}(0.6)$ & $IID_{100}$    \\ \midrule % \hline
                         
\multirow{7}{*}{\rotatebox{90}{CIFAR100}}& FedAvg  & 40.70$\pm$0.24 & 42.85$\pm$0.26 & 44.37$\pm$0.40 & 38.17$\pm$0.32	& 40.19$\pm$0.26 & 42.19$\pm$0.52  \\
                         & FedProx                 & 40.39$\pm$0.23 & 42.51$\pm$0.34 & 44.21$\pm$0.64 & 38.27$\pm$0.46 & 39.90$\pm$0.42 & 42.24$\pm$0.98  \\
                         & SCAFFOLD                & 29.36$\pm$0.39 & 33.30$\pm$0.48 & 37.96$\pm$0.26 & 23.25$\pm$0.54 & 29.98$\pm$0.19 & 32.77$\pm$0.25  \\
                         & FedPRV                  & 38.35$\pm$1.11 & 42.91$\pm$0.49 & 45.91$\pm$0.05 & 30.65$\pm$0.74 & 36.58$\pm$0.14 & 39.96$\pm$0.30     \\
                         & q-FedAvg                & 40.34$\pm$0.60 & 42.61$\pm$0.63 & 44.43$\pm$0.28 & 38.15$\pm$0.48 & 40.20$\pm$0.10 & 42.04$\pm$0.65      \\
                         & FedMix                  &\underline{42.51$\pm$0.28} &\underline{44.16$\pm$0.26} &\underline{45.65$\pm$0.31}
                                                   &\underline{39.78$\pm$0.07} &\underline{41.43$\pm$0.84} &\underline{43.63$\pm$0.64} \\
                         & FedCFA                  & \textbf{46.96$\pm$1.04} & \textbf{49.32$\pm$0.20} & \textbf{48.31$\pm$0.53} 
                                                   & \textbf{46.71$\pm$0.59} & \textbf{49.18$\pm$0.75} & \textbf{47.86$\pm$1.22} \\ \midrule %\hline

\multirow{7}{*}{\rotatebox{90}{CIFAR10}}& FedAvg   &65.88$\pm$0.32  &73.95$\pm$0.16   &75.43$\pm$0.51  &62.87$\pm$0.12  &70.99$\pm$0.70  &72.82$\pm$0.35   \\
                         & FedProx                 &72.23$\pm$0.44  &77.68$\pm$0.03 &76.93$\pm$0.28  &70.36$\pm$0.75  &75.47$\pm$0.53  &73.36$\pm$0.47   \\
                         & SCAFFOLD                &33.05$\pm$5.57  & 54.58$\pm$3.95 &75.96$\pm$0.57	 &34.69$\pm$2.16  &56.17$\pm$1.91  &71.84$\pm$0.77  \\
                         & FedPRV                  & 59.42$\pm$2.26	& 71.52$\pm$0.21 & 77.42$\pm$0.04 & 55.43$\pm$1.74 & 67.11$\pm$0.76 & 76.16$\pm$0.37   \\
                         & q-FedAvg                & 71.71$\pm$1.05	& 77.96$\pm$0.19 & 76.92$\pm$0.09 & 70.04$\pm$1.55 & 75.47$\pm$0.52 & 73.68$\pm$0.33   \\
                         & FedMix                  & \underline{74.61$\pm$0.74}	& \underline{78.64$\pm$0.53} & \underline{77.90$\pm$0.17} 
                                                   & \underline{73.91$\pm$0.79} & \underline{77.11$\pm$0.31} & \underline{73.93$\pm$0.06}  \\
                         & FedCFA                  & \textbf{75.89$\pm$1.00} & \textbf{82.43$\pm$0.08} & \textbf{83.36$\pm$0.51} 
                                                   & \textbf{75.76$\pm$0.15} & \textbf{81.73$\pm$0.12} & \textbf{81.68$\pm$0.89} \\  
% \midrule  % \hline
% \multirow{7}{*}{\rotatebox{90}{Fashion-MNIST}}& FedAvg   & 91.52$\pm$0.04 & 92.16$\pm$0.12 & 92.37$\pm$0.12 & 91.63$\pm$0.28	& 92.19$\pm$0.09 & 92.02$\pm$0.13      \\
%                          & FedProx                 & 91.39$\pm$0.10 & 92.42$\pm$0.07 & 92.38$\pm$0.10 & 91.57$\pm$0.05 & 92.20$\pm$0.13 & 92.06$\pm$0.30      \\
%                          & SCAFFOLD                & 87.65$\pm$0.13 & 90.07$\pm$0.19 & 91.81$\pm$0.14 & 87.81$\pm$0.10 & 90.00$\pm$0.02 & 91.09$\pm$0.05       \\
%                          & FedPRV                  & 89.57$\pm$0.05 & 91.42$\pm$0.12 & 92.60$\pm$0.28 & 89.01$\pm$0.17 & 91.12$\pm$0.05 & 92.27$\pm$0.08      \\
%                          & q-FedAvg                & 91.43$\pm$0.15 & 92.27$\pm$0.14 & 92.26$\pm$0.14 & 91.53$\pm$0.04 & 92.19$\pm$0.18 & 91.99$\pm$0.11      \\
%                          & FedMix                  & \textbf{91.76$\pm$0.13}    & \underline{92.51$\pm$0.04} &\underline{92.84$\pm$0.13}
%                                                    & \underline{91.88$\pm$0.13} & \underline{92.54$\pm$0.11} &\underline{92.30$\pm$0.11}   \\
%                          & FedCFA                  & \underline{91.75$\pm$0.05} & \textbf{92.86$\pm$0.31}    & \textbf{92.86$\pm$0.07} 
%                                                    & \textbf{91.90$\pm$0.05}    & \textbf{92.90$\pm$0.23}    & \textbf{92.32$\pm$0.22} \\ %\hline
                        
                         % \toprule[1pt]
\bottomrule
\end{tabular}
\caption{The top-1 accuracy (\%) after running 500 communication rounds using different methods on CIFAR100, CIFAR10. }
\label{tab:Main-Experiment}
\end{table*}

\subsection{Main Comparison}
We adopt two metrics, the global model accuracy after 500 rounds and the number of communication rounds required to achieve target accuracy, to evaluate the performance of FedCFA. %our proposed FedCFA.
Compared with six baseline methods on six datasets, FedCFA shows obvious advantages.
We set the target accuracy on CIFAR10, and compare rounds needed for different FL methods. 
Results in Table \ref{tab:Main-Experiment-2} show most other methods fail to achieve FedCFA's target accuracy within 1000 rounds, illustrating superior performance and efficiency of FedCFA in FL. %distributed FL.
In addition, we analyze the effect of FedCFA on improving model accuracy under different experimental settings. The experimental results are shown in Table \ref{tab:Main-Experiment} and \ref{tab:Main-Experiment-1-2}.

\vpara{Different Dataset.} We verify the effectiveness of FedCFA on six datasets. 
FEMNIST and Sent140 are relatively simple, and various algorithms can achieve high accuracy. 
FedCFA has only a slight advantage over other methods. On CIFAR10, CIFAR100 and Tiny-ImageNet, FedCFA shows obvious superiority. 
These three datasets feature more complex images and greater inter-client data distribution disparities.
In this case, FedCFA can effectively mitigate data heterogeneity in FL and improve global model accuracy.
For example, under the $Dir_{100}(0.6)$ partition of CIFAR100, 
FedCFA outperforms the top baseline by 7.75\%.

\begin{table}[t]
\footnotesize
\centering
\setlength{\tabcolsep}{0.3em}
\begin{tabular}{cccccc}
\toprule
\multirow{2}{*}{Method} & \multicolumn{2}{c}{Tiny-ImageNet}               & \multicolumn{1}{c}{FEMNIST}     & \multicolumn{1}{c}{Sent140} \\ 
\cmidrule(lr){2-3}\cmidrule(lr){4-4}\cmidrule(lr){5-5}
                        % & $\alpha$=0.2   & $\alpha$=0.6   & IID            & Non-IID            & Non-IID        & IID           \\ \midrule % \hline
                        &$Dir_{60}(0.2)$ &$Dir_{60}(0.6)$  & Non-IID            & Non-IID                 \\ \midrule
FedAvg                  & 27.39$\pm$0.13 & 30.90$\pm$0.29  
                        & 81.31$\pm$0.94 & \underline{68.10$\pm$0.48}  \\
FedProx                 & 27.34$\pm$0.05 & 30.78$\pm$0.16  & 81.63$\pm$0.08 & \underline{68.10$\pm$0.44} \\
q-FedAvg                & 26.89$\pm$0.07 & 30.70$\pm$0.13 
                        & 81.90$\pm$0.58 & 68.04$\pm$0.71  \\
FedMix                  & \underline{28.01$\pm$0.19}    & \underline{32.43$\pm$0.13}
                        & \underline{82.31$\pm$0.21}	& 67.97$\pm$0.39 \\
FedCFA                  & \textbf{30.70$\pm$0.68} & \textbf{32.86$\pm$0.77} 
                        & \textbf{83.19$\pm$0.54} & \textbf{69.26$\pm$0.37}   \\ \bottomrule % \hline
\end{tabular}
\caption{The top-1 accuracy (\%) after running 500 communication rounds on Tiny-ImageNet, FEMNIST, Sent140.}
\label{tab:Main-Experiment-1-2}
% \end{table}
% \begin{table}[t]
\footnotesize
\centering
\setlength{\tabcolsep}{0.28em}
\begin{tabular}{ccccccc}
\toprule
Method                  &$Dir_{60}(0.2)$  & $IID_{60}$      &$Dir_{100}(0.2)$ & $IID_{100}$    \\ \midrule
Target                  & 75.5           & 78.4           & 73.6           & 75.6   \\ \midrule % \hline
FedAvg                  & ($>$,66.22)    & ($>$,74.99)   & ($>$,62.95)    & ($>$,72.42)   \\
FedProx                 & ($>$,74.03)    & ($>$,77.16)    & ($>$,72.37)     & ($>$,73.46)   \\
SCAFFOLD                & ($>$,32.77)    & \underline{(693,78.46)} & ($>$,38.34) & \underline{(800,75.68)}   \\
q-FedAvg                & ($>$,72.23)     & ($>$,77.04)    & ($>$,72.03)    & ($>$,73.56)   \\
FedMix                  &\underline{(610,75.69)}  & ($>$,77.79) & ($>$,72.25)    & ($>$,74.00)   \\
FedCFA       & \textbf{(375,75.58)} & \textbf{(453,78.41)} & \textbf{(427,73.61)} & \textbf{(408,75.67)} \\ \bottomrule %\hline
\end{tabular}
\caption{Communication rounds required by each algorithm to achieve the target accuracy. (R, Acc) denotes the result, where R is the communication round and Acc is the model accuracy (\%). If an algorithm cannot achieve the target accuracy within 1000 rounds, we use "$>$" to denotes R, and use the highest accuracy within 1000 rounds as Acc.}
\label{tab:Main-Experiment-2}
\end{table}

\begin{figure}[t]
\centering
    \includegraphics[width=0.65\linewidth]{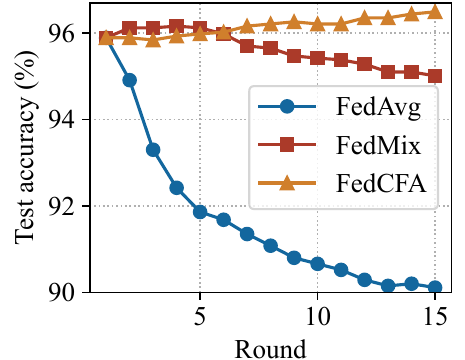}
    \caption{The top-1 accuracy (\%) on MNIST data with Simpson's Paradox.}
    \label{fig:mnist_color_sp_data}
\end{figure}

\vpara{Different Data Heterogeneity.} In view of the heterogeneity and imbalance of client data, we model the heterogeneity of label distribution $P(Y)$, the heterogeneity of feature distribution $P(X)$, and the heterogeneity of conditional feature distribution $P(X|Y)$ among clients.
We also compare the performance of each algorithm under IID partition. Through counterfactual transformation, FedCFA can narrow the gap between local and global data. 
Consequently, as Table \ref{tab:Main-Experiment} and \ref{tab:Main-Experiment-1-2} illustrate, FedCFA enhances global model accuracy under both Non-IID and IID partition.

\vpara{Different Number of Clients.} 
We consider client counts of 60 and 100.
As Table \ref{tab:Main-Experiment} shows, all baseline accuracies notably drop with increasing client numbers due to intensified data distribution heterogeneity.
However, FedCFA is able to maintain high model accuracy. 
This reveals FedCFA can effectively adapt to varying client numbers and has strong robustness and resistance to data heterogeneity.

\vpara{Data with Simpson’s Paradox.} We built a dataset with Simpson's Paradox based on MNIST, by coloring the numbers 1 and 7 and distributing them to 5 clients according to the color depth. For each client, the color of number 1 is darker than that of number 7. For example, in client 1, number 1 is yellow and number 7 is white. This data division makes the data of each client biased, and the model may learn false color-label relationship. We pre-trained an MLP to 95\% accuracy, used it as the initial weight for FL, and conducted experiments using FedCFA, FedAvg, and FedMix.
The results in Figure \ref{fig:mnist_color_sp_data} show that FedAvg and FedMix are affected by Simpson's paradox and their accuracy decreases. FedCFA eliminates false feature-label associations through counterfactual transformation, generates counterfactual samples to make local distribution close to global distribution, and improves model accuracy.

\begin{figure}[t]
\centering
    \includegraphics[width=0.76\linewidth]{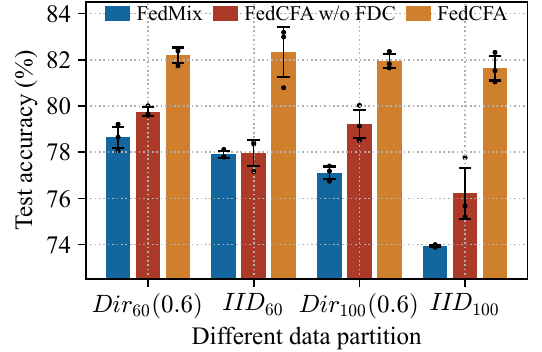}
    \caption{Error Bar Chart: At experimental settings of different random seeds and data partition.
    }
    \label{fig:error_bar}
% \end{figure}
% \begin{figure}[t]
\centering
    \includegraphics[width=0.75\linewidth]{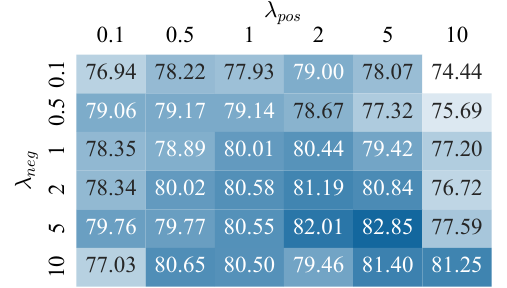}
    \caption{The top-1 accuracy (\%) of FedCFA with different proportions of $\lambda_{neg}$ and $\lambda_{pos}$.}
    \label{fig:hotmap}
\end{figure}

\subsection{Analysis of FedCFA}
This section explores the impact of hyperparameters on FedCFA by comparing FedCFA performance under different hyperparameter settings on CIFAR10.

\vpara{Factor Decorrelation Module.} We set $\lambda_{corr}$ to 0.1 and 
compare FedCFA's performance with and without FDC under CIFAR10's four data partition methods.
As Figure \ref{fig:error_bar} illustrates, FDC regularization enhances the improvement effect of the counterfactual module on model accuracy.	

\vpara{Ablation Study of Counterfactual Modules.} We conduct ablation study on two counterfactual modules under $Dir_{60}(0.6)$ setting to verify their efficacy. The two counterfactual modules are used to generate counterfactual positive and negative samples respectively. We compare four cases: FedCFA without counterfactual modules, with only positive, only negative, and with both counterfactual modules.
Table \ref{tab:Ablation-Experiment} reveals that counterfactual module can improve FedCFA performance. However, if only one counterfactual module is used, the model cannot fully utilize the contrastive learning between positive and negative samples, and the accuracy improvement is limited.
Collaborative operation of two counterfactual modules can markedly boost FedCFA's accuracy, surpassing the additive effect of individual module actions.

\vpara{Proportion of Counterfactual Modules Losses.} 
To explore the impact of the counterfactual module loss ratio on model performance, this paper conducts experiments under $Dir_{60}(0.6)$ setting.
We use a set of six different coefficients $R=\{0.1, 0.5, 1, 2, 5, 10\}$, where $\lambda_{neg},\lambda_{pos} \in R$, to obtain 36 different loss combinations. 
Figure \ref{fig:hotmap} indicates steady model performance enhancement as $\lambda_{neg}$ and $\lambda_{pos}$ rise simultaneously.
However, non-synchronous increases of $\lambda_{neg}$ and $\lambda_{pos}$ lead to unbalanced contrastive learning on counterfactual positive and negative samples, causing unstable model performance gains.
Moreover, we find excessively high counterfactual module loss ratios impair model's learning on original data distributions, reducing model accuracy.

\vpara{Application Position of Counterfactuals in Models.} 
As shown in Table \ref{tab:hyperparameter-Experiment}, Hook represents the layer at which the counterfactual is applied in the ResNet18 model. 
The smaller the Hook, the closer the counterfactual is to the input layer of the model. To explore the impact of applying counterfactuals at different layers on model performance, we conduct counterfactual experiments on different layers of ResNet18. We found that applying counterfactuals in intermediate layers achieves the best results. 
This is because at these levels, the model can not only extract key factors, but also avoid losing too many dispensable factors through excessive transformation of the data, making the generated counterfactual positive samples of low quality.

\begin{table}[t]
\centering
\footnotesize
\setlength{\tabcolsep}{0.5em}
\begin{tabular}{c|cccc}
\toprule
FedCFA   & Cls.  & Cls.+Pos. & Cls.+Neg. & Cls.+Pos.+Neg.  \\ \midrule
Accuracy & 78.17 & 78.86     & 78.87     & \textbf{80.01} \\ 
\bottomrule
\end{tabular}
\caption{Ablation study of counterfactual modules. Cls. denotes $\mathcal{L}_{cls}$, Pos. denotes $\mathcal{L}_{pos}$, and Neg. denotes $\mathcal{L}_{neg}$.}
\label{tab:Ablation-Experiment}
% \end{table}
% \begin{table}[t]
\footnotesize
    \centering
    \setlength{\tabcolsep}{0.3em}
    \renewcommand{\arraystretch}{0.8}
    \footnotesize
    \begin{tabular}{ccccccccc}
    \toprule
                           & \multicolumn{4}{c}{$Dir_{60}(0.6)$}                 & \multicolumn{4}{c}{$Dir_{100}(0.6)$}                \\ \midrule %\hline
    \multirow{2}{*}{Hook}  & 0     & 1              & 2              & 4     & 0              & 1     & 2              & 4     \\ \cmidrule(r){2-5} \cmidrule(r){6-9} 
                           & 77.65 & 77.87          & \textbf{80.01} & 75.31 & 75.89          & 75.96 & \textbf{80.03}  & 73.54 \\ \midrule % \hline
    \multirow{2}{*}{Topk}  & 8     & 16             & 24             & 32    & 8              & 16    & 24             & 32    \\ \cmidrule(r){2-5} \cmidrule(r){6-9} 
                           & 78.53 & 79.62          & \textbf{80.01} & 78.57 & 76.51          & 78.57 & \textbf{80.03} & 79.01 \\ \bottomrule % \hline
    \end{tabular}
\captionof{table}{The top-1 accuracy (\%) of FedCFA under different hyperparameter settings.}
    \label{tab:hyperparameter-Experiment}
\end{table}

\vpara{Number of Factors for Counterfactual Transformation.} In order to explore the effect of counterfactual module, we use different Topk values to control the degree of counterfactual transformation, that is, how many dispensable/indispensable factors are selected for counterfactual transformation. We select four Topk values in $\{8, 16, 24, 32\}$ and compare the models accuracy. 
As shown in Table \ref{tab:hyperparameter-Experiment}, as Topk increases, model performance first increases and then decreases. When Topk is 24, the model can achieve optimal performance. This is because when topk is too small, FedCFA cannot select all dispensable/indispensable factors. When topk is too large, the selected factors are not pure enough. Both cases lead to reduce counterfactual samples quality and affect the contrastive learning effect of model.

\section{Conclusion}
\label{sec:Conclusions}
This paper proposes FedCFA, a novel FL framework leveraging counterfactual learning.
FedCFA effectively alleviates Simpson's Paradox impacts by integrating counterfactual samples into local model training, aligning local data distributions with the global.
Additionally, FedCFA uses a factor decorrelation loss to decouple and constrain different factors in the data, ensuring the quality of counterfactual samples. We conduct extensive experiments on six datasets and demonstrate that FedCFA outperforms existing FL methods in terms of both efficiency and model accuracy.
For future research, we can explore other correlation analysis techniques to improve the framework, especially methods that can capture non-linear correlation between factors to extract more independent factors.

\section{Acknowledgments}
This work was supported by the National Science and Technology Major Project (2022ZD0119100), the National Natural Science Foundation of China (No. 62402429, 62441605), the Key Research and Development Program of Zhejiang Province (No. 2024C03270). Additionally, this work was partially supported by ZJU Kunpeng\&Ascend Center of Excellence, Ningbo Yongjiang Talent Introduction Programme (2023A-397-G).

\bibliography{arxiv}

\begin{thebibliography}{44}
\providecommand{\natexlab}[1]{#1}

\bibitem[{Alfeo et~al.(2023)Alfeo, Zippo, Catrambone, Cimino, Toschi, and Valenza}]{alfeo2023local}
Alfeo, A.~L.; Zippo, A.~G.; Catrambone, V.; Cimino, M.~G.; Toschi, N.; and Valenza, G. 2023.
\newblock From local counterfactuals to global feature importance: efficient, robust, and model-agnostic explanations for brain connectivity networks.
\newblock \emph{Computer Methods and Programs in Biomedicine}, 236: 107550.

\bibitem[{Caldas et~al.(2018)Caldas, Duddu, Wu, Li, Kone{\v{c}}n{\`y}, McMahan, Smith, and Talwalkar}]{caldas2018leaf}
Caldas, S.; Duddu, S. M.~K.; Wu, P.; Li, T.; Kone{\v{c}}n{\`y}, J.; McMahan, H.~B.; Smith, V.; and Talwalkar, A. 2018.
\newblock Leaf: A benchmark for federated settings.
\newblock \emph{arXiv preprint arXiv:1812.01097}.

\bibitem[{Cao and Gong(2022)}]{cao2022mpaf}
Cao, X.; and Gong, N.~Z. 2022.
\newblock Mpaf: Model poisoning attacks to federated learning based on fake clients.
\newblock In \emph{Proceedings of the IEEE/CVF Conference on Computer Vision and Pattern Recognition}, 3396--3404.

\bibitem[{Chai, Liu, and Yang(2022)}]{chai2022cross}
Chai, B.; Liu, K.; and Yang, R. 2022.
\newblock Cross-Domain Federated Data Modeling on Non-IID Data.
\newblock \emph{Computational Intelligence and Neuroscience}, 2022.

\bibitem[{Chen and Vikalo(2023)}]{chen2023federated}
Chen, H.; and Vikalo, H. 2023.
\newblock Federated learning in non-iid settings aided by differentially private synthetic data.
\newblock In \emph{Proceedings of the IEEE/CVF Conference on Computer Vision and Pattern Recognition}, 5026--5035.

\bibitem[{Cheng et~al.(2022)Cheng, Wang, Zhang, and Cheng}]{cheng2022differentially}
Cheng, A.; Wang, P.; Zhang, X.~S.; and Cheng, J. 2022.
\newblock Differentially private federated learning with local regularization and sparsification.
\newblock In \emph{Proceedings of the IEEE/CVF Conference on Computer Vision and Pattern Recognition}, 10122--10131.

\bibitem[{Chow et~al.(2023)Chow, Liu, Wei, Ilhan, and Wu}]{chow2023stdlens}
Chow, K.-H.; Liu, L.; Wei, W.; Ilhan, F.; and Wu, Y. 2023.
\newblock STDLens: Model Hijacking-resilient Federated Learning for Object Detection.
\newblock In \emph{Proceedings of the IEEE/CVF Conference on Computer Vision and Pattern Recognition}, 16343--16351.

\bibitem[{Fang and Ye(2022)}]{fang2022robust}
Fang, X.; and Ye, M. 2022.
\newblock Robust federated learning with noisy and heterogeneous clients.
\newblock In \emph{Proceedings of the IEEE/CVF Conference on Computer Vision and Pattern Recognition}, 10072--10081.

\bibitem[{Gao et~al.(2022)Gao, Fu, Li, Chen, Xu, and Xu}]{gao2022feddc}
Gao, L.; Fu, H.; Li, L.; Chen, Y.; Xu, M.; and Xu, C.-Z. 2022.
\newblock Feddc: Federated learning with non-iid data via local drift decoupling and correction.
\newblock In \emph{Proceedings of the IEEE/CVF conference on computer vision and pattern recognition}, 10112--10121.

\bibitem[{Go, Bhayani, and Huang(2009)}]{go2009twitter}
Go, A.; Bhayani, R.; and Huang, L. 2009.
\newblock Twitter sentiment classification using distant supervision.
\newblock \emph{CS224N project report, Stanford}, 1(12): 2009.

\bibitem[{Hamman and Dutta(2024)}]{hamman2024demystifying}
Hamman, F.; and Dutta, S. 2024.
\newblock Demystifying Local \& Global Fairness Trade-offs in Federated Learning Using Partial Information Decomposition.
\newblock In \emph{The Twelfth International Conference on Learning Representations}.

\bibitem[{Hao et~al.(2021)Hao, El-Khamy, Lee, Zhang, Liang, Chen, and Duke}]{hao2021towards}
Hao, W.; El-Khamy, M.; Lee, J.; Zhang, J.; Liang, K.~J.; Chen, C.; and Duke, L.~C. 2021.
\newblock Towards fair federated learning with zero-shot data augmentation.
\newblock In \emph{Proceedings of the IEEE/CVF Conference on Computer Vision and Pattern Recognition}, 3310--3319.

\bibitem[{Ilhan, Su, and Liu(2023)}]{ilhan2023scalefl}
Ilhan, F.; Su, G.; and Liu, L. 2023.
\newblock ScaleFL: Resource-Adaptive Federated Learning With Heterogeneous Clients.
\newblock In \emph{Proceedings of the IEEE/CVF Conference on Computer Vision and Pattern Recognition}, 24532--24541.

\bibitem[{Karimireddy et~al.(2020)Karimireddy, Kale, Mohri, Reddi, Stich, and Suresh}]{karimireddy2020scaffold}
Karimireddy, S.~P.; Kale, S.; Mohri, M.; Reddi, S.; Stich, S.; and Suresh, A.~T. 2020.
\newblock Scaffold: Stochastic controlled averaging for federated learning.
\newblock In \emph{International conference on machine learning}, 5132--5143. PMLR.

\bibitem[{Kone{\v{c}}n{\`y} et~al.(2016)Kone{\v{c}}n{\`y}, McMahan, Yu, Richt{\'a}rik, Suresh, and Bacon}]{konevcny2016federated}
Kone{\v{c}}n{\`y}, J.; McMahan, H.~B.; Yu, F.~X.; Richt{\'a}rik, P.; Suresh, A.~T.; and Bacon, D. 2016.
\newblock Federated learning: Strategies for improving communication efficiency.
\newblock \emph{arXiv preprint arXiv:1610.05492}.

\bibitem[{Krizhevsky, Hinton et~al.(2009)}]{krizhevsky2009learning}
Krizhevsky, A.; Hinton, G.; et~al. 2009.
\newblock Learning multiple layers of features from tiny images.

\bibitem[{Li et~al.(2021{\natexlab{a}})Li, Sun, Zeng, Zhang, Li, and Chen}]{li2021fedmask}
Li, A.; Sun, J.; Zeng, X.; Zhang, M.; Li, H.; and Chen, Y. 2021{\natexlab{a}}.
\newblock Fedmask: Joint computation and communication-efficient personalized federated learning via heterogeneous masking.
\newblock In \emph{Proceedings of the 19th ACM Conference on Embedded Networked Sensor Systems}, 42--55.

\bibitem[{Li et~al.(2023{\natexlab{a}})Li, Esfandiari, Schmidt, Alstr{\o}m, and Stich}]{li2023synthetic}
Li, B.; Esfandiari, Y.; Schmidt, M.~N.; Alstr{\o}m, T.~S.; and Stich, S.~U. 2023{\natexlab{a}}.
\newblock Synthetic data shuffling accelerates the convergence of federated learning under data heterogeneity.
\newblock \emph{arXiv preprint arXiv:2306.13263}.

\bibitem[{Li et~al.(2023{\natexlab{b}})Li, Schmidt, Alstr{\o}m, and Stich}]{li2023effectiveness}
Li, B.; Schmidt, M.~N.; Alstr{\o}m, T.~S.; and Stich, S.~U. 2023{\natexlab{b}}.
\newblock On the effectiveness of partial variance reduction in federated learning with heterogeneous data.
\newblock In \emph{Proceedings of the IEEE/CVF Conference on Computer Vision and Pattern Recognition}, 3964--3973.

\bibitem[{Li, He, and Song(2021)}]{li2021model}
Li, Q.; He, B.; and Song, D. 2021.
\newblock Model-contrastive federated learning.
\newblock In \emph{Proceedings of the IEEE/CVF conference on computer vision and pattern recognition}, 10713--10722.

\bibitem[{Li et~al.(2021{\natexlab{b}})Li, Hu, Beirami, and Smith}]{li2021ditto}
Li, T.; Hu, S.; Beirami, A.; and Smith, V. 2021{\natexlab{b}}.
\newblock Ditto: Fair and robust federated learning through personalization.
\newblock In \emph{International Conference on Machine Learning}, 6357--6368. PMLR.

\bibitem[{Li et~al.(2020{\natexlab{a}})Li, Sahu, Talwalkar, and Smith}]{liTi2020federated}
Li, T.; Sahu, A.~K.; Talwalkar, A.; and Smith, V. 2020{\natexlab{a}}.
\newblock Federated learning: Challenges, methods, and future directions.
\newblock \emph{IEEE signal processing magazine}, 37(3): 50--60.

\bibitem[{Li et~al.(2020{\natexlab{b}})Li, Sahu, Zaheer, Sanjabi, Talwalkar, and Smith}]{li2020federated}
Li, T.; Sahu, A.~K.; Zaheer, M.; Sanjabi, M.; Talwalkar, A.; and Smith, V. 2020{\natexlab{b}}.
\newblock Federated optimization in heterogeneous networks.
\newblock \emph{Proceedings of Machine learning and systems}, 2: 429--450.

\bibitem[{Li et~al.(2019)Li, Sanjabi, Beirami, and Smith}]{li2019fair}
Li, T.; Sanjabi, M.; Beirami, A.; and Smith, V. 2019.
\newblock Fair resource allocation in federated learning.
\newblock \emph{arXiv preprint arXiv:1905.10497}.

\bibitem[{Li et~al.(2022)Li, Zhang, Liu, and Liu}]{li2022auditing}
Li, Z.; Zhang, J.; Liu, L.; and Liu, J. 2022.
\newblock Auditing privacy defenses in federated learning via generative gradient leakage.
\newblock In \emph{Proceedings of the IEEE/CVF Conference on Computer Vision and Pattern Recognition}, 10132--10142.

\bibitem[{Liao et~al.(2023)Liao, Gao, Zhao, and Xu}]{liao2023adaptive}
Liao, D.; Gao, X.; Zhao, Y.; and Xu, C.-Z. 2023.
\newblock Adaptive Channel Sparsity for Federated Learning Under System Heterogeneity.
\newblock In \emph{Proceedings of the IEEE/CVF Conference on Computer Vision and Pattern Recognition}, 20432--20441.

\bibitem[{Luo et~al.(2023)Luo, Li, Lan, and Gao}]{luo2023gradma}
Luo, K.; Li, X.; Lan, Y.; and Gao, M. 2023.
\newblock GradMA: A Gradient-Memory-based Accelerated Federated Learning with Alleviated Catastrophic Forgetting.
\newblock In \emph{Proceedings of the IEEE/CVF Conference on Computer Vision and Pattern Recognition}, 3708--3717.

\bibitem[{McMahan et~al.(2017)McMahan, Moore, Ramage, Hampson, and y~Arcas}]{mcmahan2017communication}
McMahan, B.; Moore, E.; Ramage, D.; Hampson, S.; and y~Arcas, B.~A. 2017.
\newblock Communication-efficient learning of deep networks from decentralized data.
\newblock In \emph{Artificial intelligence and statistics}, 1273--1282. PMLR.

\bibitem[{Mendieta et~al.(2022)Mendieta, Yang, Wang, Lee, Ding, and Chen}]{mendieta2022local}
Mendieta, M.; Yang, T.; Wang, P.; Lee, M.; Ding, Z.; and Chen, C. 2022.
\newblock Local learning matters: Rethinking data heterogeneity in federated learning.
\newblock In \emph{Proceedings of the IEEE/CVF Conference on Computer Vision and Pattern Recognition}, 8397--8406.

\bibitem[{Mou et~al.(2021)Mou, Geng, Welten, Rong, Decker, and Beyan}]{mou2021optimized}
Mou, Y.; Geng, J.; Welten, S.; Rong, C.; Decker, S.; and Beyan, O. 2021.
\newblock Optimized federated learning on class-biased distributed data sources.
\newblock In \emph{Joint European Conference on Machine Learning and Knowledge Discovery in Databases}, 146--158. Springer.

\bibitem[{Ovi et~al.(2023)Ovi, Dey, Roy, and Gangopadhyay}]{ovi2023mixed}
Ovi, P.~R.; Dey, E.; Roy, N.; and Gangopadhyay, A. 2023.
\newblock Mixed Quantization Enabled Federated Learning to Tackle Gradient Inversion Attacks.
\newblock In \emph{Proceedings of the IEEE/CVF Conference on Computer Vision and Pattern Recognition}, 5045--5053.

\bibitem[{Qu et~al.(2022)Qu, Zhou, Liang, Xia, Wang, Adeli, Fei-Fei, and Rubin}]{qu2022rethinking}
Qu, L.; Zhou, Y.; Liang, P.~P.; Xia, Y.; Wang, F.; Adeli, E.; Fei-Fei, L.; and Rubin, D. 2022.
\newblock Rethinking architecture design for tackling data heterogeneity in federated learning.
\newblock In \emph{Proceedings of the IEEE/CVF Conference on Computer Vision and Pattern Recognition}, 10061--10071.

\bibitem[{Reguieg et~al.(2023)Reguieg, Hanjri, Kamili, and Kobbane}]{reguieg2023comparative}
Reguieg, H.; Hanjri, M.~E.; Kamili, M.~E.; and Kobbane, A. 2023.
\newblock A Comparative Evaluation of FedAvg and Per-FedAvg Algorithms for Dirichlet Distributed Heterogeneous Data.
\newblock \emph{arXiv preprint arXiv:2309.01275}.

\bibitem[{Rothchild et~al.(2020)Rothchild, Panda, Ullah, Ivkin, Stoica, Braverman, Gonzalez, and Arora}]{rothchild2020fetchsgd}
Rothchild, D.; Panda, A.; Ullah, E.; Ivkin, N.; Stoica, I.; Braverman, V.; Gonzalez, J.; and Arora, R. 2020.
\newblock Fetchsgd: Communication-efficient federated learning with sketching.
\newblock In \emph{International Conference on Machine Learning}, 8253--8265. PMLR.

\bibitem[{Seo and Elmroth(2023)}]{seo2023madfed}
Seo, E.; and Elmroth, E. 2023.
\newblock MadFed: Enhancing Federated Learning with Marginal-data Model Fusion.
\newblock \emph{IEEE Access}.

\bibitem[{Xie and Song(2023)}]{xie2023fedkl}
Xie, Z.; and Song, S. 2023.
\newblock Fedkl: Tackling data heterogeneity in federated reinforcement learning by penalizing kl divergence.
\newblock \emph{IEEE Journal on Selected Areas in Communications}, 41(4): 1227--1242.

\bibitem[{Yoon et~al.(2021)Yoon, Shin, Hwang, and Yang}]{yoon2021fedmix}
Yoon, T.; Shin, S.; Hwang, S.~J.; and Yang, E. 2021.
\newblock Fedmix: Approximation of mixup under mean augmented federated learning.
\newblock \emph{arXiv preprint arXiv:2107.00233}.

\bibitem[{Zeng et~al.(2023)Zeng, Liang, Hu, Wang, and Xu}]{JMLR:v24:22-0440}
Zeng, D.; Liang, S.; Hu, X.; Wang, H.; and Xu, Z. 2023.
\newblock FedLab: A Flexible Federated Learning Framework.
\newblock \emph{Journal of Machine Learning Research}, 24(100): 1--7.

\bibitem[{Zhang et~al.(2021)Zhang, Xie, Bai, Yu, Li, and Gao}]{zhang2021survey}
Zhang, C.; Xie, Y.; Bai, H.; Yu, B.; Li, W.; and Gao, Y. 2021.
\newblock A survey on federated learning.
\newblock \emph{Knowledge-Based Systems}, 216: 106775.

\bibitem[{Zhang et~al.(2020)Zhang, Jiang, Wang, Kuang, Zhao, Zhu, Yu, Yang, and Wu}]{zhang2020devlbert}
Zhang, S.; Jiang, T.; Wang, T.; Kuang, K.; Zhao, Z.; Zhu, J.; Yu, J.; Yang, H.; and Wu, F. 2020.
\newblock Devlbert: Learning deconfounded visio-linguistic representations.
\newblock In \emph{Proceedings of the 28th ACM International Conference on Multimedia}, 4373--4382.

\bibitem[{Zhang et~al.(2024)Zhang, Miao, Nie, Li, Chen, Feng, Kuang, and Wu}]{zhang2024transferring}
Zhang, S.; Miao, Q.; Nie, P.; Li, M.; Chen, Z.; Feng, F.; Kuang, K.; and Wu, F. 2024.
\newblock Transferring Causal Mechanism over Meta-representations for Target-Unknown Cross-domain Recommendation.
\newblock \emph{ACM Transactions on Information Systems}, 42(4): 1--27.

\bibitem[{Zhao et~al.(2023)Zhao, Elkordy, Sharma, Ezzeldin, Avestimehr, and Bagchi}]{zhao2023resource}
Zhao, J.~C.; Elkordy, A.~R.; Sharma, A.; Ezzeldin, Y.~H.; Avestimehr, S.; and Bagchi, S. 2023.
\newblock The Resource Problem of Using Linear Layer Leakage Attack in Federated Learning.
\newblock In \emph{Proceedings of the IEEE/CVF Conference on Computer Vision and Pattern Recognition}, 3974--3983.

\bibitem[{Zhu et~al.(2021)Zhu, Xu, Liu, and Jin}]{zhu2021federated}
Zhu, H.; Xu, J.; Liu, S.; and Jin, Y. 2021.
\newblock Federated learning on non-IID data: A survey.
\newblock \emph{Neurocomputing}, 465: 371--390.

\bibitem[{Zhu, Ma, and Blaschko(2023)}]{zhu2023confidence}
Zhu, J.; Ma, X.; and Blaschko, M.~B. 2023.
\newblock Confidence-aware personalized federated learning via variational expectation maximization.
\newblock In \emph{Proceedings of the IEEE/CVF Conference on Computer Vision and Pattern Recognition}, 24542--24551.

\end{thebibliography}

% \bigskip

\newpage
\section{Appendix}
% \appendix
\setcounter{table}{0}
\setcounter{figure}{0}
\renewcommand{\thetable}{\Roman{table}}
\renewcommand{\thefigure}{\Roman{figure}}

% \newpage

% \twocolumn[
% \begin{@twocolumnfalse}
% 	\section*{\centering{FedCFA: Alleviating Simpson's Paradox in Model Aggregation \\ with Counterfactual Federated Learning \\ Appendix \\[25pt]}}
%     % \section*{\centering{Appendix\\[25pt]}}	% 标题
% \end{@twocolumnfalse}
% ]

% \section{Data Partition Visualization}
\subsection{Data Partition Visualization}
\label{sec:Data-Partition-Visualization}

\begin{figure*}[!hbp]
  \centering
   \includegraphics[width=1.0\linewidth]{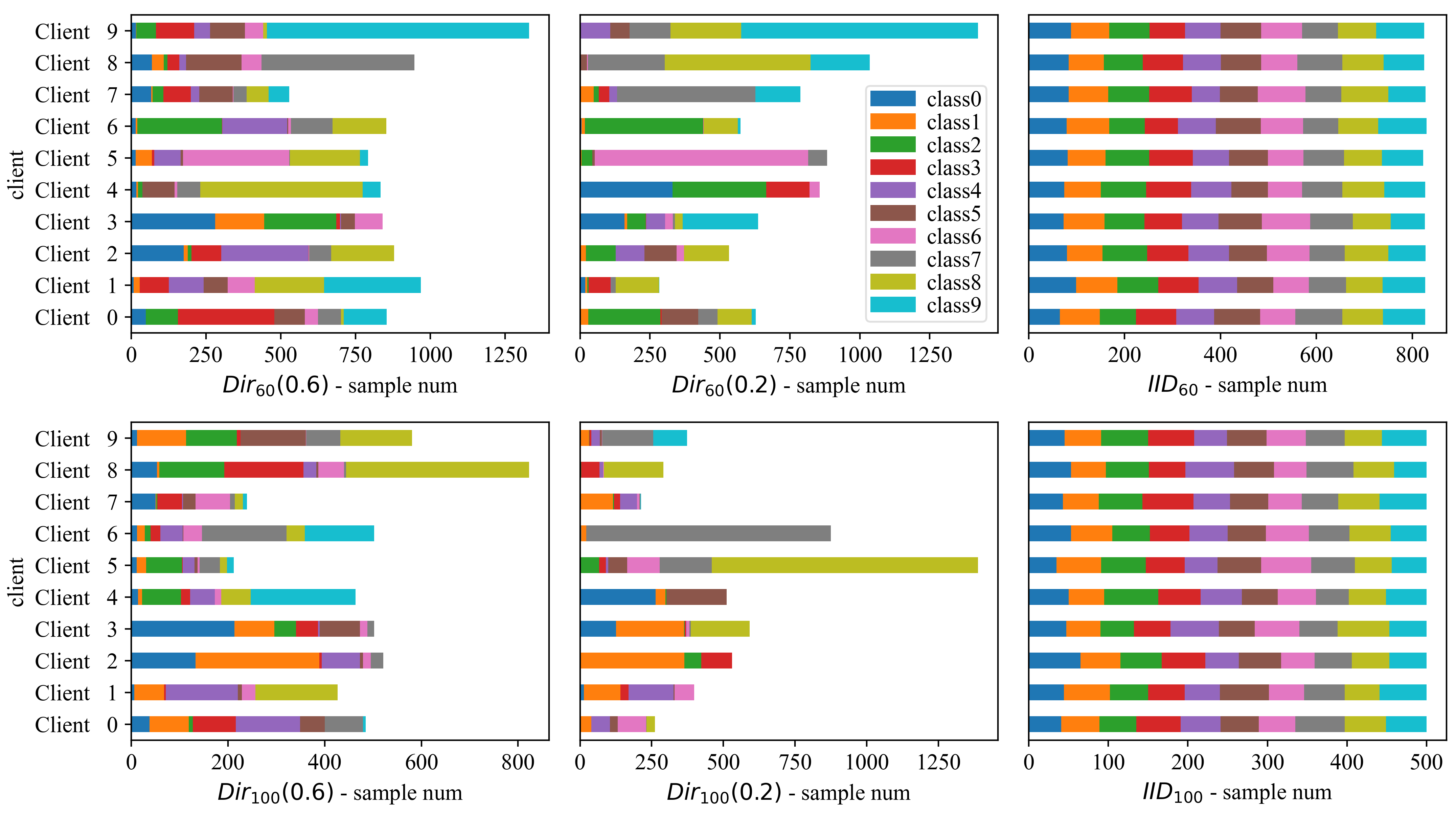}
   \caption{Distribution of data of the first 10 clients in CIFAR10 partition.}
   \label{fig:data_partition}
\end{figure*}

We partition CIFAR10 through six different strategies: $Dir_{60}(0.6)$, $Dir_{60}(0.2)$, $Dir_{100}(0.2)$, $Dir_{100}(0.6)$, $IID_{60}$, $IID_{100}$, and visually present the partitioned data. To simulate Non-IID data scenarios in the real world, we employ the Dirichlet distribution. The Figure \ref{fig:data_partition} clearly shows that as the Dirichlet distribution parameter $\alpha$ decreases, the data differences between clients gradually increase. 
% We considered two different settings of the number of clients, 60 and 100, in our experiments. 
In our experiments, we consider two distinct configurations for the number of clients: 60 and 100. 
It can be clearly observed from the visualization results that as the number of clients increases, the heterogeneity of data distribution intensifies.

\begin{figure}[H]
  \centering
   \includegraphics[width=1.0\linewidth]{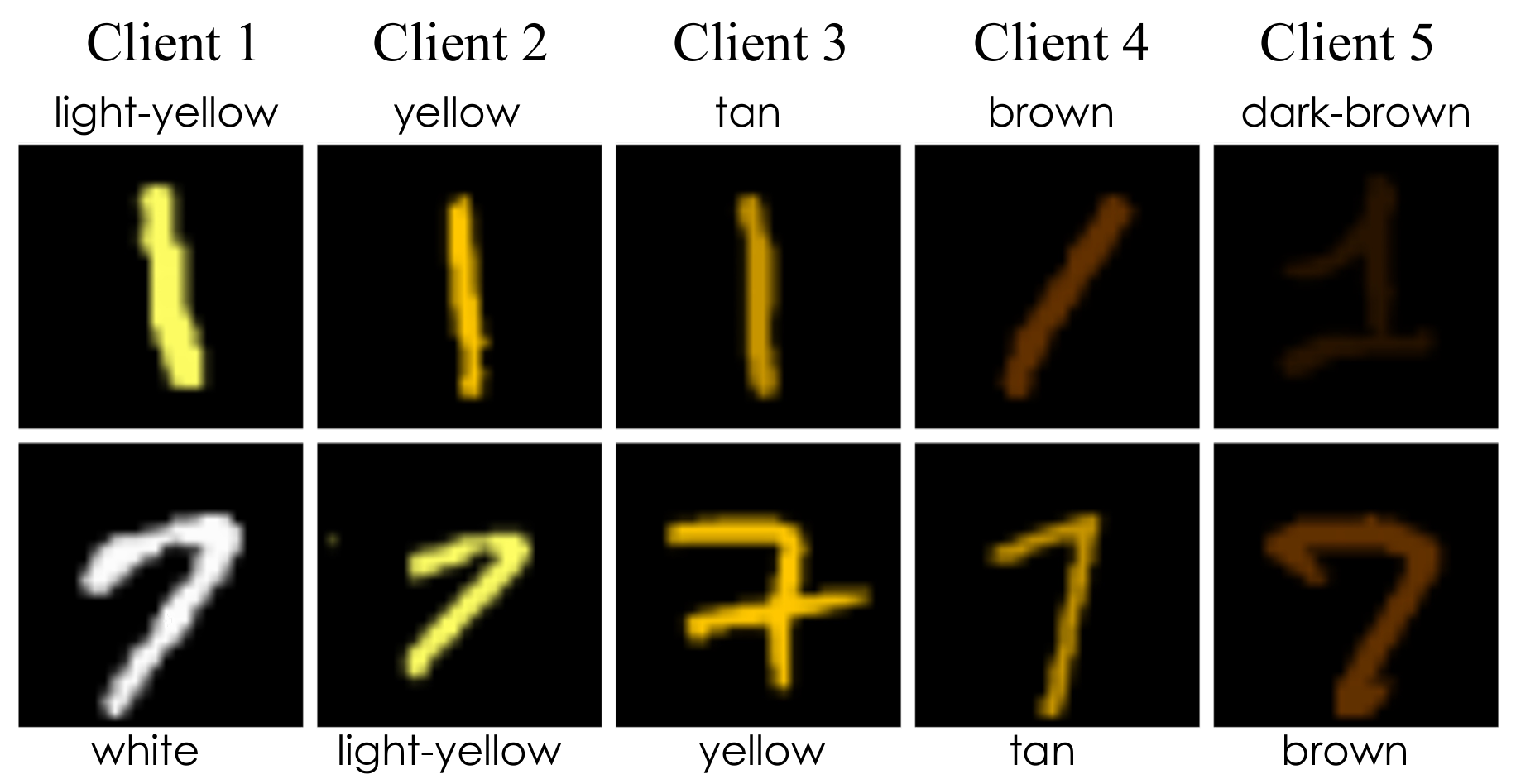}
   \caption{Data with Simpson’s Paradox.}
   \label{fig:data_with_sp}
\end{figure}

\vpara{Data with Simpson’s Paradox.} 
In our experiments, we construct a dataset with Simpson's Paradox based on MNIST. 
Specifically, we focus on the digit classes '1' and '7', applying distinct color schemes to these images and partitioning them across five clients according to color intensity. 
% Within each client's dataset, the color of the digit '1' is consistently darker than the color of the digit '7'.
In each client's dataset, the color of the digit '1' is consistently darker than that of the digit '7'. 
For instance, the dataset of client 1 contains the digit '1' in yellow and the digit '7' in white. 
% This setting effectively simulates the scenario where color intensity and label are spuriously correlated, and can be used to study Simpson’s Paradox in a federated learning setting. 
This configuration effectively simulates a scenario where color intensity and label are spuriously correlated, facilitating the study of Simpson’s Paradox in federated learning. 
The handwritten digit image data divided into each client is shown in Figure \ref{fig:data_with_sp}.


\begin{figure*}[t]
  \centering
   \includegraphics[width=1.0\linewidth]{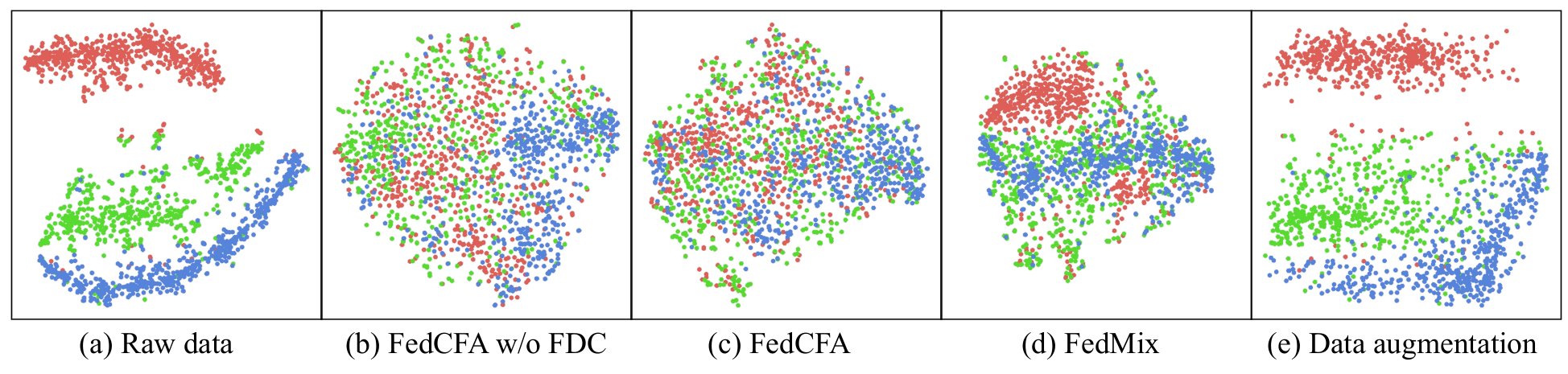}
   \caption{T-SNE visualization of factor distribution of CIFAR10.}
   \label{fig:t-sne}
\end{figure*}

\begin{table*}[t]
\small
\centering
\setlength{\tabcolsep}{1em}
\begin{tabular}{cccccccc}
\toprule
                         & Method                  &$Dir_{60}(0.2)$ &$Dir_{60}(0.6)$ & $IID_{60}$      &$Dir_{100}(0.2)$ &$Dir_{100}(0.6)$ & $IID_{100}$    \\ \midrule % \hline

\multirow{7}{*}{\rotatebox{90}{Fashion-MNIST}}& FedAvg   & 91.52$\pm$0.04 & 92.16$\pm$0.12 & 92.37$\pm$0.12 & 91.63$\pm$0.28	& 92.19$\pm$0.09 & 92.02$\pm$0.13      \\
                         & FedProx                 & 91.39$\pm$0.10 & 92.42$\pm$0.07 & 92.38$\pm$0.10 & 91.57$\pm$0.05 & 92.20$\pm$0.13 & 92.06$\pm$0.30      \\
                         & SCAFFOLD                & 87.65$\pm$0.13 & 90.07$\pm$0.19 & 91.81$\pm$0.14 & 87.81$\pm$0.10 & 90.00$\pm$0.02 & 91.09$\pm$0.05       \\
                         & FedPRV                  & 89.57$\pm$0.05 & 91.42$\pm$0.12 & 92.60$\pm$0.28 & 89.01$\pm$0.17 & 91.12$\pm$0.05 & 92.27$\pm$0.08      \\
                         & q-FedAvg                & 91.43$\pm$0.15 & 92.27$\pm$0.14 & 92.26$\pm$0.14 & 91.53$\pm$0.04 & 92.19$\pm$0.18 & 91.99$\pm$0.11      \\
                         & FedMix                  & \textbf{91.76$\pm$0.13}    & \underline{92.51$\pm$0.04} &\underline{92.84$\pm$0.13}
                                                   & \underline{91.88$\pm$0.13} & \underline{92.54$\pm$0.11} &\underline{92.30$\pm$0.11}   \\
                         & FedCFA                  & \underline{91.75$\pm$0.05} & \textbf{92.86$\pm$0.31}    & \textbf{92.86$\pm$0.07} 
                                                   & \textbf{91.90$\pm$0.05}    & \textbf{92.90$\pm$0.23}    & \textbf{92.32$\pm$0.22} \\ %\hline
                        
\bottomrule
\end{tabular}

\caption{The top-1 accuracy (\%) after running 500 communication rounds using different methods on Fashion-MNIST. }
\label{tab:appendix-Main-Experiment}
% \end{table*}
% \begin{table*}[t]
\footnotesize
\centering
\setlength{\tabcolsep}{1em}
\renewcommand{\arraystretch}{1.2}
\begin{tabular}{cccccccc}
\hline
\multirow{3}{*}{Method} & \multicolumn{3}{c}{Tiny-ImageNet}               & \multicolumn{2}{c}{FEMNIST}     & \multicolumn{2}{c}{Sent140} \\ \cmidrule(r){2-4}\cmidrule(r){5-6}\cmidrule(r){7-8}
                        & $Dir_{60}(0.2)$   & $Dir_{60}(0.6)$   & $IID_{60}$            & Non-IID        & IID            & Non-IID        & IID           \\ \hline
FedAvg                  & 27.39$\pm$0.13 & 30.90$\pm$0.29 & \underline{33.01$\pm$0.57} 
                        & 81.31$\pm$0.94 & 84.82$\pm$0.14 & \underline{68.10$\pm$0.48} & {78.90$\pm$0.09} \\
FedProx                 & 27.34$\pm$0.05 & 30.78$\pm$0.16 & 32.93$\pm$0.52 & 81.63$\pm$0.08	& 85.00$\pm$0.13 & \underline{68.10$\pm$0.44} & \underline{78.97$\pm$0.25} \\
q-FedAvg                & 26.89$\pm$0.07 & 30.70$\pm$0.13 & 32.75$\pm$0.55 
                        & 81.90$\pm$0.58 & 84.69$\pm$0.01 & {68.04$\pm$0.71} & {78.90$\pm$0.09} \\
FedMix                  & \underline{28.01$\pm$0.19} & \underline{32.43$\pm$0.13}    & 32.45$\pm$0.78 
                        & \underline{82.31$\pm$0.21}	& \textbf{85.29$\pm$0.06} & 67.97$\pm$0.39 & 78.66$\pm$0.11 \\
FedCFA                  & \textbf{30.70$\pm$0.68}  & \textbf{32.86$\pm$0.77}    & \textbf{33.57$\pm$1.09} 
                        & \textbf{83.19$\pm$0.54} & \underline{85.19$\pm$0.17}  & \textbf{69.26$\pm$0.37}  & \textbf{78.98$\pm$0.00} \\ \hline
\end{tabular}
\caption{The top-1 accuracy (\%) after running 500 communication rounds using different methods on Tiny-ImageNet, FEMNIST, Sent140.}
\label{tab:appendix-Main-Experiment-1-2}
\end{table*}

\subsection{T-SNE Visualization}
We employ t-SNE to visualize the factor distributions of the CIFAR10 under the $Dir_{100}(0.2)$ partition. 
% As illustrated in Figure \ref{fig:t-sne}, the distribution of the factors for local data across three clients reveals notable disparities compared to the global distribution, with distinct colors representing each client.
As illustrated in Figure \ref{fig:t-sne}, the distribution of the factors for local data across three clients reveals notable disparities compared to the global distribution, where each color represents a client. 
This visualization reflects the presence of Simpson's Paradox in real-world Non-IID datasets. 

As shown in Figure \ref{fig:t-sne}(a), the original data distributions exhibit heterogeneity among the clients. As seen in Figure \ref{fig:t-sne}(d), after applying the Mixup operation of FedMix, the data distributions among the three clients become more closely, but distinct clusters are still observable, indicating that inter-client data heterogeneity persists.
% Additionally, we also try employing augmentation techniques such as color jittering and random flipping. 
Moreover, we also explore augmentation strategies including color jittering and random flipping. 
As evidenced in Figure \ref{fig:t-sne}(e), these augmentations result in a slight generalization of the clients' local data distributions. 
However, these naive augmentations do not mitigate the Simpson's Paradox. 
% This may stem from the inability of these augmentations to selectively process specific features across the dataset, leading to treatment across all features, or potentially due to the presence of complex, latent features that require encoding for effective extraction.
This could be attributed to the lack of selective processing capabilities of these augmentations for specific features within the dataset, resulting in indiscriminate treatment of all features, or possibly owing to the existence of intricate, latent features necessitating specialized encoding for efficient extraction.
Notably, Figure \ref{fig:t-sne}(b) and (c) demonstrate that the counterfactual transformations introduced by FedCFA enable the local data distributions of the three clients to align more closely with the global distribution, with FDC effectively facilitating this harmonization.

\subsection{Supplementary Explanation of FedCFA}
In the initial iteration of the FedCFA algorithm, the server has not yet generated global average data, so the global average data sent to clients is None. At this stage, the client does not activate the counterfactual transformation module, but calculates local average data for updating the global average data on the server. 

For the factor decorrelation (FDC) module, when the batchsize of the data is 2, the Pearson correlation coefficients between factors are all equal to 1. Therefore, we stipulate that when the batchsize is less than 3, the FDC module is not activated.

\subsection{Additional Experiments}
In this study, we conduct extensive experiments under more experimental settings to verify the superiority of the FedCFA over other existing methods. As presented in Table \ref{tab:appendix-Main-Experiment}, our experiments on the Fashion-MNIST dataset reveal that, although the overall accuracy of all algorithms is very high due to the simplicity of the dataset, FedCFA still shows an advantage over the other algorithms. This shows that the FedCFA algorithm not only performs well in complex scenarios, but also maintains its advantage in simple settings.

For Table 2 in paper, we also conducted experiments on Tiny-imagenet, FEMNIST, and Sent140 data under IID partition, verifying that our algorithm still improves the model accuracy under IID configuration. The complete data is shown in Table \ref{tab:appendix-Main-Experiment-1-2}.

For Table 3 in paper, we also conduct experiments on CIFAR10 under $Dir_{60}(0.6)$ and $Dir_{100}(0.6)$ partitions. The complete data is shown in Table \ref{tab:appendix-Main-Experiment-2}. 
The results show that our algorithm improves the convergence speed of model on data with different degrees of heterogeneity, can achieve higher accuracy with fewer communication rounds, and reduces communication costs. 

As shown in Table \ref{tab:hyperparameter-Experiment-l}, we analyze the weight $\lambda_{corr}$ of FDC loss. We select four parameters $\{0.001, 0.01, 0.1, 1.0\}$ and conduct experiments on CIFAR10 under $Dir_{60}(0.6)$ partition. The results show that when $\lambda_{corr}=0.1$, FedCFA can achieve the best performance.

\begin{table*}[t]
\footnotesize
\centering
\setlength{\tabcolsep}{1em}
\renewcommand{\arraystretch}{1.0}
\begin{tabular}{ccccccc}
\toprule
Method                  &$Dir_{60}(0.2)$ &$Dir_{60}(0.6)$ & $IID_{60}$      &$Dir_{100}(0.2)$ &$Dir_{100}(0.6)$ & $IID_{100}$    \\ \midrule % \hline
Target                  & 75.5           & 80.0           & 78.4           & 73.6           & 80.0           & 75.6   \\ \midrule
FedAvg                  & ($>$,66.22)    & ($>$,73.76)     & ($>$,74.99)   & ($>$,62.95)    & ($>$,71.14)     & ($>$,72.42)   \\
FedProx                 & ($>$,74.03)    & ($>$,78.84)    & ($>$,77.16)    & ($>$,72.37)    & ($>$,76.15)    & ($>$,73.46)   \\
SCAFFOLD                & ($>$,32.77)    & ($>$,64.26)    & \underline{(693,78.46)} & ($>$,38.34) & ($>$,64.75)  & \underline{(800,75.68)}   \\
q-FedAvg                & ($>$,72.23)    & ($>$,78.52)   & ($>$,77.04)    & ($>$,72.03)    & ($>$,76.05)    & ($>$,73.56)   \\
FedMix                  &\underline{(610,75.69)} & ($>$,79.86) & ($>$,77.79) & ($>$,72.25)   & ($>$,77.92)  & ($>$,74.00)   \\
FedCFA       & \textbf{(375,75.58)} & \textbf{(468,80.01)} & \textbf{(453,78.41)} & \textbf{(427,73.61)} & \textbf{(449,80.03)} & \textbf{(408,75.67)} \\ \bottomrule
\end{tabular}
\caption{The number of communication rounds required by each algorithm to achieve the target accuracy. (R, Acc) denotes the result, where R is the communication round and Acc is the model accuracy (\%). If an algorithm cannot achieve the target accuracy within 1000 rounds, we use "$>$" to denotes R, and use the highest accuracy within 1000 rounds as Acc.}
\label{tab:appendix-Main-Experiment-2}
\end{table*}
\begin{table}[t]
\centering
\footnotesize
\setlength{\tabcolsep}{1em}
\renewcommand{\arraystretch}{1.2}\
\begin{tabular}{c|cccc}
\toprule
$\lambda_{corr}$  & 0.001  & 0.01     & 0.1     & 1.0  \\ \midrule
Accuracy          & 81.44  & 81.49    & \textbf{82.85}   & 80.67 \\ \bottomrule
\end{tabular}
\caption{Analysis of hyperparameter $\lambda_{corr}$.}
\label{tab:hyperparameter-Experiment-l}
\end{table}

\end{document}